# Framework for Question-Answering in Sanskrit through Automated Construction of Knowledge Graphs

**Hrishikesh Terdalkar**  **Arnab Bhattacharya**
`hrishirt@cse.iitk.ac.in`  `arnabb@cse.iitk.ac.in`
Dept. of Computer Science and Engineering,
Indian Institute of Technology Kanpur,
India.

## Abstract

Sanskrit (saṃskṛta) enjoys one of the largest and most varied literature in the whole world. Extracting the knowledge from it, however, is a challenging task due to multiple reasons including complexity of the language and paucity of standard natural language processing tools. In this paper, we target the problem of building knowledge graphs for particular types of relationships from saṃskṛta texts. We build a natural language question-answering system in saṃskṛta that uses the knowledge graph to answer factoid questions. We design a framework for the overall system and implement two separate instances of the system on human relationships from mahābhārata and rāmāyaṇa, and one instance on synonymous relationships from bhāvaprakāśa nighaṇṭu, a technical text from āyurveda. We show that about 50% of the factoid questions can be answered correctly by the system. More importantly, we analyse the shortcomings of the system in detail for each step, and discuss the possible ways forward.

## 1 Introduction and Motivation

Sanskrit (IAST[1]: saṃskṛta, Devanagari: संस्कृत) is one of the most ancient and richest languages in the world. Its literature boasts of text spanning every facet of life and contains works on mathematics, arts, sciences, religion, philosophy, etc. Unfortunately, the large volume of such works and the relative lack of proficiency in the language have kept treasures in those text hidden from the common man. Unraveling information from these texts in a targeted and systematic manner can not only help in enhancing the knowledge systems but can also revive an interest in the language.

Many of these texts are technical in nature, prime examples of which include āyurveda (आयुर्वेद) texts such as bhāvaprakāśa (भावप्रकाश). The nighaṇṭu (निघण्टु) portion of bhāvaprakāśa is compiled as a glossary of the various substances (dravya, द्रव्य) and their properties (guṇa, गुण). Although the information is generally provided in a format that enables scholars to study and analyse it systematically, the large volume of such texts makes it harder for any individual to extract all the information. An automated system can, therefore, greatly aid this processing of information. However, to the best of our knowledge, there does not exist any system that can query this knowledge trove directly and automatically.

While it can be argued that English translations of bhāvaprakāśa nighaṇṭu are available, and building information retrieval (IR) systems for it is a routine for today's IR/NLP tools, there are two main shortcomings of it. First, there are many such nighaṇṭu texts and translations in English are available for only a minuscule number of them. Second, and more importantly, many of the translations of saṃskṛta texts had been done without a proper understanding of the context and culture in which they were composed in the first place. They may had been forced to use English words and phrases that are not a true reflection of the spirit of the original meaning. A notable case in point, as mentioned by Swami Vivekananda himself, is the word śraddhā (श्रद्धा), for which the English translation "regards" is not enough.

---

[1]Entire paper uses the IAST encoding scheme for writing Sanskrit words in romanized format. `https://en.wikipedia.org/wiki/International_Alphabet_of_Sanskrit_Transliteration`

Thus, it is always best to rely on the original language. The need of the hour, hence, is to use natural language processing (NLP) of saṃskṛta itself to understand the texts in saṃskṛta.

Our aim in this work is to take the first step towards a concrete NLP task, namely, natural language question-answering in saṃskṛta. In particular, we aim to design a *framework* that processes saṃskṛta texts, extracts the information in it, and stores it in a format that can be queried using questions posed in saṃskṛta.

We propose to store the knowledge base (KB) in a knowledge graph (KG) format. KGs have a rich structure and store the information in the form of entities (as nodes) and relationships (as edges between the nodes). The edges are directed, and both the nodes and edges can store labels describing their attributes. There are multiple off-the-shelf tools available for storing and querying KGs, including graph databases[2], Property Graphs[3], Resource Description Framework (RDF) (Lassila et al. (1998)), Gremlin queries[4], SPARQL queries[5], etc. The popularity of KBs such as YAGO (Suchanek et al. (2007)), DBpedia (Auer et al. (2007)) and Freebase (Bollacker et al. (2008)) is a testament to their success.

We also propose question-answering as a concrete example of the use of such KGs and a way of measuring the effectiveness of the system. Various online question-answering fora such as Quora[6] and quizzes serve as a motivation. We particularly choose the two epics of India, namely, mahābhārata and rāmāyaṇa, categorized as itihāsa in saṃskṛta literature, and questions on human relationships within them, as examples for our framework due to their popularity and ease of establishment of the ground truth. We also work with bhāvaprakāśa nighaṇṭu to highlight the usage for technical texts.

The framework brings to the fore multiple challenges. First, the state of the art of natural language processing in Indian languages, unfortunately, is not as advanced as that in English or some other European languages. Indian languages, and in particular saṃskṛta, are morphologically richer. Therefore, tasks such as lemmatization and parts-of-speech tagging are harder and more error-prone in these languages. Second, some technical texts use their own jargon where certain words may be used in a specific meaning. For example, aṣṭādhyāyī, a work on saṃskṛta grammar by pāṇini uses specific combinations of grammatical cases (vibhakti) to denote which action is to be performed.[7] Third, names in saṃskṛta are meaningful words and, therefore, identifying named entities is particularly hard. An extremely interesting example in rāmāyaṇa is janaka (जनक), which means "father" in general, but is also the name of a prominent character. Fourth, synonyms are often used to refer to the same person. In many cases, higher-order grammar rules are required to parse the meaning of a word and understand that it is a synonym. For example, it is not mentioned anywhere in the rāmāyaṇa text that dāśarathī is the son of daśaratha and, hence, synonymous to rāma. However, saṃskṛta grammar rules make it obvious to someone who understands the language. Unfortunately, automatic language processing tools are incapable of using such higher-order rules at present.

Nair and Kulkarni (2010) have proposed a model for extracting implicit knowledge from amarakośa and storing it in a structured manner, and have constructed a tool for answering queries using this knowledge. Kulkarni et al. (2010) have built a Sanskrit WordNet[8] by expanding the

---

[2] https://en.wikipedia.org/wiki/Graph_database
[3] https://en.wikipedia.org/wiki/Graph_database#Labeled-property_graph
[4] https://docs.janusgraph.org/latest/gremlin.html
[5] https://www.w3.org/TR/rdf-sparql-query/
[6] https://www.quora.com
[7] The presence of nominative (prathamā), genitive (ṣaṣṭhī) and locative (saptamī) cases in the same sentence might not convey any special meaning in a normal text, but, in aṣṭādhyāyī, it specifies a process to be followed to transform words, e.g., rule 6.1.77 from aṣṭādhyāyī (iko yaṇaci, इको यणचि) contains words ikaḥ (ṣaṣṭhī), yaṇ (prathamā), aci (saptamī), which is to be interpreted as "an इक् letter which is followed by an अच् letter is converted to a corresponding यण् letter".
[8] http://www.cfilt.iitb.ac.in/wordnet/webswn/english_version.php

Hindi WordNet. A production grammar for human relationships in saṃskṛta was proposed in Bhargava and Lambek (1992). It works for solitary words and cannot be directly used for text. Automatic translation tools, if available, can also be used where the entire text is translated to English and the KG is built from the translated text. However, we could not find any such tools. Although Sanskrit-English dictionaries[9] provide a word-level translation of selected words from saṃskṛta to English, word-level translation often does not produce meaningful or grammatically correct text. We, thus, decided to use only the text as available in saṃskṛta. In future, we will explore the use of such tools and methods.

The rest of the paper is organized as follows. In Section 2, we explain the generic framework of the question-answering system. There exist some excellent tools for saṃskṛta that aid us in the analysis. For other cases, we build our own heuristic rule-based systems. In Section 3, we describe the automatic construction of the knowledge graph while the details of the various modules of the system are described in Section 4. Since bhāvaprakāśa nighaṇṭu is a technical text, we highlight its specialized processing in Section 5. In Section 6, we analyse the results of our experiments. Finally, in Section 7, we discuss the lessons learnt and future directions.

## 2 Proposed Framework

### 2.1 Knowledge Graphs (KG)

Knowledge graphs (KG) model real-world entities as nodes. Relationships among the entities are modelled as (directed) edges. For example, in a KG about human relationships in mahābhārata, arjuna and abhimanyu are nodes. They are connected by a directed edge from arjuna to abhimanyu labelled by the relationship "has-son" (putra).

In English, there have been several efforts in automated KG construction, notable among them being YAGO, DBpedia, Freebase, etc. Suchanek et al. (2007) built the YAGO ontology by crawling the Wikipedia and uniting it with WordNet using a combination of both rule-based as well as heuristic methods. Auer et al. (2007) built DBpedia that extracts knowledge present in a structured form on Wikipedia by template detection using pattern matching coupled with post-processing for quality improvement. Bollacker et al. (2008) designed Freebase, a database of tuples that is created, edited and maintained in a collaborative manner. Unfortunately, however, none of the above techniques are applicable for automatically building knowledge graphs in saṃskṛta.

Processing of text for YAGO depends on many IR/NLP tools that are available only in English and a handful of other languages, mostly European. The state of the art of these tools in saṃskṛta is still not standardized and may not be directly useful. Sanskrit Wikipedia[10] also is not as resourceful as its counterpart in English. Hence, the amount of structured information available there is minuscule compared to the vast saṃskṛta literature that is developed over several millennia. Thus, a system such as DBpedia is not possible. A collaborative effort such as Freebase is also ruled out due to a paucity of active saṃskṛta users adept in digital technologies. To the best of our knowledge, there is no work that directly builds a knowledge graph from saṃskṛta texts.

### 2.2 Triplets

A common way of encoding the relationship information is in the form of *semantic triplets*. A triplet has the structure `[subject, predicate, object]` which indicates that the entity `subject` has the relationship `predicate` with the entity `object`. Hence, the fact that arjuna has a son abhimanyu is encoded as the triplet [arjuna, has-son (putra), abhimanyu] ([अर्जुन, पुत्र, अभिमन्यु]).

The KG is built automatically by extracting such triplets from the text. We target KGs on specific types of relationships, namely, human relationships for epics, and synonymous relation-ships in nighaṇṭu. One of the foremost jobs, therefore, is to identify the relationship words. This is a corpus-independent set and depends only on the language. However, since the text is free-flowing (except in technical texts where there is a structure) and almost always written in poetry in the form of śloka, even when a relationship word is identified, the subject and object words may be anywhere around it (both before and after). śloka (श्लोक) is a semantic unit in saṃskṛta and is equivalent to a *verse*. Sometimes, one or both of these entities may not be even in the same śloka. Hence, a *context* window around the relationship word must be defined and searched for the relevant entities. Specifying the length of such a context window is not easy; if it is too short, relationships may be missed, while if it is too long, too many spurious relationships may be inferred. Even identifying the śloka boundaries may not always be trivial. Fortunately, however, these boundaries are clearly marked in the texts that we have worked on.

---

[9]https://www.sanskrit-lexicon.uni-koeln.de/
[10]https://sa.wikipedia.org/wiki

The details of how such triplets are extracted are explained in Section 3. The knowledge graph is maintained in an RDF format as a set of all such extracted triplets.

### 2.3 Questions

The next important task in the pipeline is to parse the natural language question. Since the question is also in saṃskṛta, we adopt similar processing as the text to extract triplets. In this work, we assume only factoid based questions such as "Who is the son of arjuna?" (अर्जुनस्य पुत्रः कः?) The triplet extracted from the above question will be [arjuna, has-son, X] ([अर्जुन, पुत्र, किम्]).

Since saṃskṛta is quite free with word ordering, the above question may be asked in different manners, such as अर्जुनस्य पुत्रः कः? or कः अर्जुनस्य पुत्रः? or अर्जुनस्य कः पुत्रः? All of these should yield the same triplet [अर्जुन, पुत्र, किम्].

The *inverse* question may also be asked: "Who is the father of abhimanyu?" (कः अभिमन्योः पिता?) The above can be answered only if it is known that the inverse of "has-father" is the relationship "has-son". This, again, is a property of the language and must be explicitly mentioned.

Hence, we maintain a map of such inverse relationship rules. Note that it is not always one-to-one. For example, "has-mother" is also the inverse of "has-son", and "has-father" is the inverse of "has-daughter" as well. Gender information, therefore, becomes important.

We augment the initially built knowledge graph by adding appropriate inverse relationship edges. It is ensured that an inferred inverse relationship does not contradict a directly inferred relationship from the text. The details are in Section 3.4.

Even though the questions are simple and short, they may contain *multiple* triplets. For example, a question पाण्डोः पत्न्याः भ्राता कः? may be asked by someone who does not know what the relation brother-of-wife is called in saṃskṛta. This question contains two relationships, पत्नी and भ्राता. The triplet form of these relationships would be [पाण्डु, पत्नी, किम्] corresponding to the subquestion 'Who was the wife of pāṇḍu?' and [पत्नी, भ्राता, किम्] corresponding to the subquestion 'Who was the brother of wife (of pāṇḍu)?'. All of these must be extracted correctly.

Further, they must be linked properly. In the example above, we must ensure that the object of the first triplet is the subject of the second triplet, that is, the correct triplets are [पाण्डु, पत्नी, X] and [X, भ्राता, किम्]. Here, a variable is used to denote the person that satisfies both the triplets.

Once these are correctly linked, a SPARQL query pattern is formed. The SPARQL query equivalent for the above question is

```
SELECT ?A
WHERE {
    :पाण्डु :पत्नी ?X .
    ?X :भ्रातृ ?A .
}
```

This is finally directly queried against the KG, and the answer is returned. Section 4 describes in detail the intricacies of the different steps of the question-answering system.

Figure 1 describes the overall framework. The final accuracy of the system is dependent on each of the modules of the architecture. For example, if the extracting triplets component is very erroneous, then neither the KG information is captured correctly, nor is the intention of the question understood. The overall error is a cascading effect of the errors in each of the individual components. Thus, for a successful system, each component must be reasonably accurate.

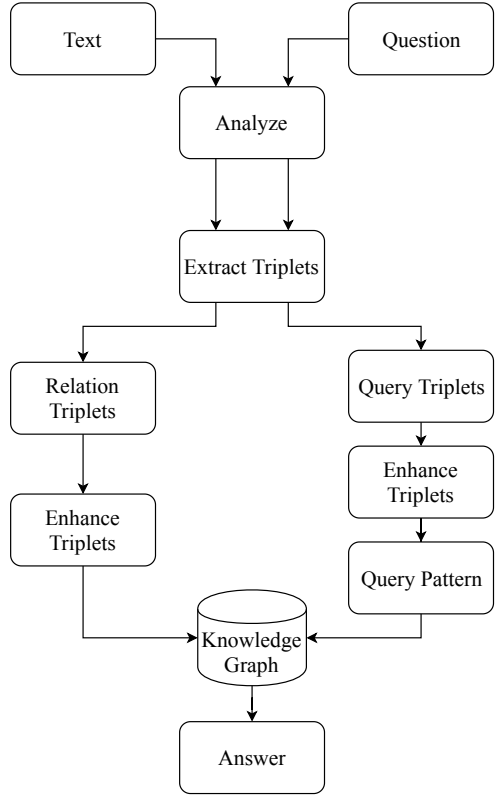

Figure 1: Overall framework of the system.

## 3 Construction of Knowledge Graph

In this section, we describe in detail the automated construction of knowledge graph (KG). The input consists of saṃskṛta text (in digital Unicode format) of an entire work (such as mahābhārata, bhāvaprakāśa nighaṇṭu, etc.) and the *type* of relationships intended (e.g., human relationships, synonymous words, etc.). The output is a set of triplets in the form [subject, predicate, object] where the predicate is of the relationship type intended and subject and object are entities. If `[a, R, b]` is an output triplet, then it implies that *object* `b` is *relation* `R` of *subject* `a`.

### 3.1 Pre-Processing of Text

saṃskṛta is a morphologically rich language. A single word root, called prātipadika (प्रातिपदिक), can yield many forms depending on the case, gender and number. Similarly, a single verb root, called dhātu (धातु), can lead to many forms as well depending on the tense, person and number. In addition, various prefixes (upasarga, उपसर्ग) and suffixes (pratyaya, प्रत्यय) get affixed to these forms to generate thousands of other forms.

Further, words are very often joined together to form compound words using either pronunciation rules through a process called sandhi (सन्धि) or semantic rules through a process called samāsa (समास). Often, both are invoked together, and a series of words are joined together to form one big compound word.

Splitting these compound words into their base words is a highly complicated procedure and may not always be unambiguous. For this step, we make use of the *Sanskrit Sandhi and*

*Compound Splitter*, a tool[11] by Hellwig and Nehrdich (2018). For example, if the input text is कर्णार्जुनयोः को श्रेष्ठः the output is कर्ण-अर्जुनयोः कः श्रेष्ठः.

The next task is to semantically analyze the *form* of the word. Again, we use a third-party *analyser tool*, *The Sanskrit Reader Companion*[12] from *The Sanskrit Heritage Platform* by Goyal et al. (2012). This tool outputs the case (**vibhakti**, विभक्ति), number (**vacana**, वचन) and gender (**liṅga**, लिङ्ग) for each word. The tool uses various abbreviations[13] to convey the linguistic information.

For the running example, the analysis yields

कर्ण ['voc.', 'sg.', 'm.']
अर्जुन ['loc.', 'du.', 'm.']
किम् ['nom.', 'sg.', 'm.']
श्रेष्ठ ['nom.', 'sg.', 'm.']

Here, 'nom.', 'loc.' and 'voc.' are abbreviations used to denote nominative case (प्रथमा), locative case (सप्तमी) and vocative case (सम्बोधन) respectively. Similarly, 'sg.' and 'du.' indicate singular and dual number (एकवचन and द्विवचन). While 'm.' denotes the masculine gender (पुंलिङ्ग).

The word श्रेष्ठ gets correctly analysed: it is in the nominative case, is in singular number, and masculine gender. However, the other words require some more adjustments. For example, the word अर्जुन is shown to be in dual number. This is output since the original compound word consisted of two persons. However, now that they are separated, it should no longer be in dual number, but adjusted to be in singular number. Similarly, the case analysis for कर्ण is wrongly output to be vocative. The reason for this again is the fact that the original structure of the compound word was lost. We adjust the case of previous words in a compound word by adopting the case of the last word in the compound word. Thus, the case for कर्ण is changed to locative, since that is the case for अर्जुन.

## 3.2 Identifying Relationship Words

Given a particular relationship type, the set of words pertaining to it is corpus-independent and is a property of the language. For example, if human relationships are targeted, in **saṃskṛta**, the (roots of the) relevant words are **pitṛ** (father, पितृ), **mātṛ** (mother, मातृ), **putra** (son, पुत्र), **putrī** (daughter, पुत्री), **pati** (husband, पति), **patnī** (wife, पत्नी), etc. Of course, these words can appear in various forms. More importantly, their synonyms can also appear. For example, all the words दुहितृ, तनया, आत्मजा mean पुत्री.

While these can be learned, since the set is mostly fixed, we have employed a key-value based approach where we have listed many of such relationship words along with their synonyms. For each such group of synonyms, there is a canonical word (e.g., पुत्री for the group of words indicating daughter) that is used in the KG.

The identification of a relationship word is simply a match from this entire set of words.

## 3.3 Identification of Triplets

Once a relationship word is identified, it forms the predicate of a triplet. The next task, therefore, is to identify the subject and object corresponding to it.

It is expected that the subject and object entities will not be too far off from the predicate word. To bound the sphere of influence or context, we use śloka (श्लोक) boundaries. Each śloka considered as a semantic unit and is akin to a verse. Fortunately, for the texts we have used, the śloka *boundaries* are clearly marked. In this work, we restrict the context to be *one* śloka before and after the one where the predicate is found, i.e., a total of 3 śloka.

Since subjects and objects are entities, they generally occur as nouns in a language. The analyser tool (*The Sanskrit Reader Companion*) described earlier marks the parts-of-speech tags of words. It, however, does not distinguish between nouns, pronouns and adjectives. Since there is a fixed set of pronouns for saṃskṛta, we use that set to correct some of the nouns. We, however, fail to distinguish the adjectives from the nouns in a satisfactory and consistent manner. This is a major future work.

---

[11] https://github.com/OliverHellwig/sanskrit/tree/master/papers/2018emnlp
[12] https://sanskrit.inria.fr/DICO/reader.fr.html
[13] All the abbreviations used by the tool are listed at https://sanskrit.inria.fr/abrevs.pdf.

Within the nouns (and adjectives), we look for those that are in the *genitive* case (षष्ठी विभक्ति). The genitive case pertains to the ṣaṣṭhī vibhakti (genitive case) and denotes sambandha (सम्बन्ध). The word sambandha in saṃskṛta literally means relationship and, therefore, a noun exhibiting genitive case is the most likely candidate for a subject. For example, the अर्जुनस्य पुत्रः अभिमन्युः आसीत् means abhimanyu was son *of* arjuna. Here, 'of arjuna' is expressed by the genitive case of the word (अर्जुन), i.e., अर्जुनस्य. Hence, all such nouns in the genitive case are marked as subjects.

The relationship word or the predicate can be in different cases, numbers and gender, though. Since the object follows the predicate, according to saṃskṛta grammar, it must be in the same case, number and gender as the predicate. We use this rule to extract objects. To be precise, an object is a noun that exhibits the same case, number and gender as the predicate word. In the sentence अर्जुनस्य पुत्रः अभिमन्युः आसीत्, word पुत्रः is the predicate word and the word अभिमन्युः is the object and both of these words are in the nominative case (प्रथमा विभक्ति).

We insert all such extracted triplets in the KG. We assume that if an entity appears multiple times, it refers to the *same* person. The above assumption is almost always correct barring some exceptional cases.[14]

## 3.4 Enhancement of Relationships

As explained earlier (in Section 2), just the base relationships may not always be enough to answer a question. If the triplet [arjuna, has-son, abhimanyu] ([अर्जुन, पुत्र, अभिमन्यु]) is stored, the question "Who is the father of abhimanyu?" (कः अभिमन्योः पिता?) cannot be answered, even though the information is present.

To be able to answer such queries, we have enhanced the KG with inverse relationships. For example, the inverse of "has-father" is "has-son". This, again, is a property of the language and are explicitly stored.

As discussed earlier, the inverse relationships are not always one-to-one. For example, "has-mother" is also the inverse of "has-son", and "has-father" is the inverse of "has-daughter" as well. Hence, we use the gender information of the subject and the object to disambiguate.

The complication does not end here. Imagine a question "Who is maternal uncle of Nakula?" (नकुलस्य मातुलः कः). This information may not be directly stored in the KG. The relationship मातुल is a composition of मातृ and भ्रातृ. These components [नकुल, मातृ, माद्री] and [माद्री, भ्रातृ, शल्य] may be present in the KG. Again, the situation is that the KG contains the information but cannot answer the question.

To solve this, derived relations could be broken into their component base parts. Thus, "has-maternal uncle" is stored as "has-mother" and "has-brother" with an additional (possibly unnamed) node in between. In particular, from the triplet [नकुल, मातुल, शल्य], two more triplets [नकुल, मातृ, X] and [X, भ्रातृ, शल्य] could be generated. If there is already such a node X, it could be used; otherwise, a new node could be created. However, addition of such *dummy* nodes has not been explored in this work.

We achieve the same result by handling this issue at the time of querying. This is discussed in Section 4.2. We maintain a list of relationships and their possible derivations from base relationships. Once more this mapping is rarely one-to-one. For example, "brother-of" can be composed of "son-of-father" and "son-of-mother". Also, the gender must be taken care of.

A particularly interesting case is "has-ancestor" and "has-descendant". These are recursive relationships, and the depth of recursion can be anything, i.e., a 'father' is an ancestor, so is an 'ancestor-of-father', and so on. We do not handle these cases in the current work.

---

[14]karṇa was the son of kuntī, and one of the kaurava was also named karṇa.

## 4 Question-Answering

We now describe one application, that of question-answering. We assume that the questions are asked directly in saṃskṛta and are about factoids, i.e., about a single piece of information. We also assume that the questions are only about the relationships that the knowledge graph encodes. If not, the question is ignored, since clearly the KG is incapable of answering it. Further, the questions are assumed to be short and consist of a single sentence only.

The question is first pre-processed in the same manner as the text (Section 3.1). To be more precise, compound words are split using *Sanskrit Sandhi and Compound Splitter* a tool by Hellwig and Nehrdich (2018), the component words are analysed using *The Sanskrit Reader Companion* from *The Sanskrit Heritage Site*, and relationship words and nouns are identified. Next, triplets are extracted.

### 4.1 Identifying Triplets

A blank triplet is initialized. The question words are scanned one by one. For each word, it is determined if it can be a subject word, a predicate word or an object word. If the word is a noun in genitive case but is not a relationship word, then it is likely to be a subject word. The relationship words directly give the predicates. The object word is generally in the nominative case. For example, consider the question अर्जुनस्य पुत्रः कः? ("Who is the son of **arjuna**?"). Since अर्जुन is in genitive case, it is the subject. The word पुत्र is the predicate. The object is किम्. The triplet formed, therefore, is [अर्जुन, पुत्र, किम्].

Once a triplet is filled up, another new triplet is initialized. This is necessary since there may be chain questions of the form अर्जुनस्य पुत्रस्य पुत्रः कः? The triplets generated from this are [अर्जुन, पुत्र, X] and [X, पुत्र, किम्].

The process goes on till all the words in the question are processed.

At the end of this phase, the triplets thus formed are called *query triplets*.

### 4.2 Enhancing Triplets

Each query triplet is next enhanced to a set of triplets, called the *enhanced triplet set*. The rules for enhancing the relationship of a query triplet is the same as that used in processing the KG triplets. In particular, each complex relation is broken into its constituent parts and new triplets are created using the aforementioned mapping of relationships to its constituents.

Suppose, a predicate (i.e., relation) `R` can be decomposed to two base predicates `R1` and `R2`. Then, if a query triplet is of the form `[A, R, B]`, then two triplets of the form `[A, R1, X]` and `[X, R2, B]` are generated. Note that `{[A, R, B]}` and `{[A, R1, X], [X, R2, B]}` are *equivalent expressions* and either of them can return the correct answer from the KG. However, since it is not known which information is stored in the KG, *both* are used.

Thus, each query triplet $QT_i$ is replaced by its enhanced triplet set $ET_i = \{QT_i\} \cup IT_i^j$ where $IT_i^j$ is a set of triplets inferred from $QT_i$, as shown in the example below.

For the question अर्जुनस्य मातुलस्य पिता कः, we first obtain the triplets {[अर्जुन, मातुल, X], [X, पितृ, किम्]}. These triplets are then enhanced by appropriately splitting the relationship मातुल using the rule मातुल = मातृ + भ्रातृ. Here, $QT =$ [अर्जुन, मातुल, X] and $IT = $ {[अर्जुन, मातृ, Y], [Y, भ्रातृ, X]}. As a result, we get two triplet sequences for this question, {[अर्जुन, मातृ, Y], [Y, भ्रातृ, X], [X, पितृ, किम्]} and {[अर्जुन, मातुल, X], [X, पितृ, किम्]}.

### 4.3 Query Pattern

If the question contains only one query triplet, then members of its enhanced triplet set form the alternate query patterns. Suppose, however, the question contains $n$ query triplets with their corresponding $n$ enhanced triplet sets $ET_1, ET_2, \cdots, ET_n$. The Cartesian product of the elements of these sets form the *alternate query patterns*. Thus, if there are 2 enhanced sets with 2 and 3 elements in them, the total number of alternate query patterns is $2 \times 3 = 6$.

Each of these alternate query patterns are posed to the KG and answer triplets are returned. The correct field of the answer triplet is returned as the factoid answer.

We have not encountered a case where alternate query patterns return different answers. If, however, such a situation arises, a further disambiguation step (possibly using majority voting, etc.) is required.

## 5 Technical Texts

We have chosen a technical text bhāvaprakāśa which is one of the important texts from āyurveda. bhāvaprakāśa nighaṇṭu is a glossary chapter from this text, which contains detailed information about the medicinal properties of various plants, animals and minerals written in a śloka format. There are 23 adhyāya in this chapter. Being a technical text, bhāvaprakāśa nighaṇṭu has more structure than rāmāyaṇa or mahābhārata.

### 5.1 Structure

The text bhāvaprakāśa nighaṇṭu loosely adheres to the following structure.

- Substances (dravya, द्रव्य) with similar properties or from the same class occur in the same chapter. For example, all the flowers are in one chapter, all the metals are in another chapter.

- Each chapter consists of various *blocks* (sets of consecutive śloka), where each block speaks about one substance.

- Each block generally has the following internal components:
    - Synonyms of the concerned substance
    - Where that substance can be found
    - Properties of the substance. e.g., colour, smell, texture, composition and other medicinal properties
    - Differences between the different varieties of the substance

While the blocks are structured to some extent, the following deviations exist.

- The length of each block is not fixed.

- The number of synonyms of each substance are not fixed.

- The order of the components of the block varies from substance to substance to a certain extent.

- Some of the internal components may, at times, be absent such as the varieties of a substance.

Importantly, the *separation* between two consecutive blocks is not marked in the text.

These points of deviation from the pattern act as hurdles in the process of understanding and exploiting the structure of a text to extract information. Understanding the structure of a text can be a challenging task. We have taken the help of domain experts[15] to form our understanding of the structure described above.

Properties (guṇa, गुण) are of the form (`name, value`). A property value can be directly attached to a substance, or it can be attached through a `property-name`. For example, a substance is "red", or, a substance has *colour* "red".

Relationships of interest can be of a number of types. Some of them are: (`substance-1, is-synonym-of, substance-2`), (`substance, property-name, property-value`), (`substance,

---

[15] We acknowledge Dr. Sai Susarla, Dean at Maharshi Veda Vyas MIT School of Vedic Sciences, Pune, India, and his team for sharing their expertise with us.

| Words | | | Nouns | | |
|---|---|---|---|---|---|
| adhyāya 1 | adhyāya 2 | All adhyāya | adhyāya 1 | adhyāya 2 | All adhyāya |
| (च, 127) | (च, 56) | (च, 946) | (कफ, 53) | (तिक्त, 39) | (पित्त, 461) |
| (तद्, 85) | (तिक्त, 39) | (तद्, 786) | (उष्ण, 47) | (कफ, 31) | (कफ, 438) |
| (किम्, 55) | (लघु, 37) | (पित्त, 461) | (पित्त, 45) | (विष, 22) | (गुरु, 254) |
| (कफ, 53) | (कफ, 31) | (कफ, 438) | (तिक्त, 34) | (उष्ण, 21) | (उष्ण, 240) |
| (उष्ण, 47) | (तु, 24) | (तु, 394) | (वात, 32) | (पित्त, 19) | (तिक्त, 237) |
| (पित्त, 45) | (किम्, 24) | (लघु, 321) | (शूल, 29) | (कुष्ठ, 18) | (वात, 204) |
| (तु, 39) | (तद्, 22) | (वा, 278) | (कुष्ठ, 28) | (अस्र, 18) | (स्मृत, 194) |
| (तथा, 35) | (विष, 22) | (अपि, 268) | (कास, 25) | (स्मृत, 17) | (कुष्ठ, 177) |
| (अपि, 34) | (उष्ण, 21) | (किम्, 266) | (कटु, 25) | (कण्डु, 16) | (गुण, 160) |
| (तिक्त, 34) | (हृत, 20) | (गुरु, 254) | (श्वास, 24) | (कटु, 16) | (लघु, 160) |

Table 1: Top-10 most frequent words, nouns and their frequencies from bhāvaprakāśa nighaṇṭu.

| | |
|---|---|
| **Counts** | Words, Nouns, Properties, Non-Properties, Special Words, Pronouns, Verbs, Case-$i$ Nouns ($i = 1, \ldots, 8$), Number-$j$ Nouns ($j =$ singular, dual, plural) |
| **Ratio to Words** | Nouns, Properties, Non-Properties, Special Words |
| **Ratio to Nouns** | Properties, Non-Properties, Special Words, Case-$i$ Nouns ($i = 1, \ldots, 8$), Number-$j$ Nouns ($j =$ singular, dual, plural) |
| **Other Ratios** | Properties to Non-Properties, Non-Properties to Properties, Special Words to Properties, Special Words to Non-Properties |

Table 2: Features of a śloka.

`has-property`, `property-value`), (`substance`, `found-at`, `location`).
When a `property` is directly attached to a substance, we assume the relationship to be `has-property`.

We have currently focused our efforts on a single relationship in the bhāvaprakāśa nighaṇṭu, namely, `is-synonym-of`. In other words, the triplets that we are interested in are of the form (`substance-1`, `is-synonym-of`, `substance-2`). Since the predicate is same for all triplets, we choose to get rid of it and think of the problem as simply *finding pairs of synonyms*.

This task is subdivided into two tasks, (1) finding śloka that contain the synonyms, and (2) given such a śloka, finding pairs of synonyms from it.

### 5.2 Property Words

The corpus is initially pre-processed in a similar manner as described in Section 3.1. However, a next layer of processing is done to extract more information.

The set of properties is a relatively small set of words. The names and values of these properties together are called *property words*. Since the *property words* recur heavily in every block that describes a substance, they are expected to have much higher frequencies than the names of substances. We test this hypothesis by performing a frequency analysis of the top words and nouns in the entire text.

Table 1 lists the top-10 most frequent words and nouns along with their frequencies. Notice that most frequent words also contain stopwords like च, तद् etc., while the list of nouns indicates that the standard property words such as वात, पित्त, कफ have a high frequency. Following this empirical evidence, we choose the top-50 most frequent nouns as "properties". The substances are chosen from the rest of the nouns.

## 5.3 Synonym śloka Identification

Generally, the different synonyms of a substance are listed in a single śloka at the beginning of a block. A set $\{n_1, n_2, \ldots n_k\}$ of nouns is called a *synonym-group* if every $n_i$ is a synonym of every other $n_j$. Any such $(n_i, n_j)$ pair is called a *synonym-pair*. A śloka that gives information about a synonym-group or synonym-pairs is referred to as a *synonym* śloka. The first task is to identify instances of such synonym śloka.

To identify a synonym śloka automatically, we use various linguistic features of a śloka and then use them in a classifier. We create a 42-dimensional feature vector per śloka. Table 2 enlists all the features used. The features are based on counts and their ratios. Some of the notable features include number of nouns, pronouns and verbs, number of property words present in a śloka, ratios of property words to total number of words, number of words in each case (विभक्ति), and so on. The category "specials" contains adverbs, conjunctions and prepositions.

Once each śloka is converted into a 42-dimensional feature vector, various classifiers and ensemble methods are used to classify into a synonym śloka or otherwise.

## 5.4 Identifying Synonymous Nouns

Once a synonym śloka is identified, the next task is to identify the synonyms from it. Given a synonym śloka, we first exclude all the property words from it. We next consider the list of all the nouns in the śloka: $\{n_1, n_2, \ldots, n_k\}$.

We call a pair of nouns $(n_i, n_j)$ a *synonym pair* if both $n_i$ and $n_j$ have the same case (विभक्ति) as well as the same number (वचन). We do not use the gender (लिङ्ग) information since there are examples of synonymous substance names that belong to different genders. For example, चव्य (neuter), चव्यिका (feminine) and ऊषणा (feminine) form a synonym group.

## 6 Experiments and Results

In this section, we present our experiments and discuss the results. The code is written in Python3. All experiments are done on Intel(R) Core(TM) i7-4770 CPU @ 3.40GHz system with 16 GB RAM running Ubuntu 16.04.6 OS. RDF is used for storing the knowledge graph, and querying is done using SPARQL querying language. Python library RDFlib is used for working with RDF and SPARQL.

### 6.1 Datasets

We have worked with texts containing two types of relationships:

1. **Human Relationships:** The two well-known epics of ancient India, rāmāyaṇa and mahābhārata, contain numerous characters and relationships among them. We have, thus, used them as datasets for human relationships.

2. **Synonymous Relationships of Substances:** āyurveda, the traditional Indian system of medicine, has a rich source of information about medicinal plants and substances. We considered bhāvaprakāśa nighaṇṭu, a glossary chapter of the āyurveda text bhāvaprakāśa as the dataset. It enlists numerous medicinal plants and substances along with their properties and inter-relationships. In this work, we only consider the relationship "is-synonym-of".

Table 3 shows the statistics about the datasets considered.

### 6.2 Knowledge Graph from rāmāyaṇa and mahābhārata

Table 4 shows the various statistics about the knowledge graphs constructed from the datasets rāmāyaṇa and mahābhārata.

While pre-processing the text requires a large amount of time, the other steps are significantly faster. The querying times are in microseconds.

| Dataset | rāmāyaṇa | mahābhārata | bhāvaprakāśa nighaṇṭu |
|---|---|---|---|
| Type | Classical | Classical | Technical |
| Chapters | 7 (kāṇḍa) | 18 (parvan) | 23 (adhyāya) |
| Documents | 606 | 2,327 | 23 |
| śloka | 23,934 | 81,603 | 4,244 |
| Words (total) | 2,69,603 | 17,49,709 | 31,532 |
| Words (unique) | 16,083 | 55,366 | 5,976 |
| Nouns (total) | 1,52,878 | 6,36,781 | 19,689 |
| Nouns (unique) | 9,553 | 20,545 | 3,684 |

Table 3: Statistics of the various datasets used.

| | | rāmāyaṇa | mahābhārata |
|---|---|---|---|
| Time taken | Preprocessing | ∼ 3.5 days | ∼ 13 days |
| | Triplet Extraction | 14.18 sec | 57.19 sec |
| | Triplet Enhancement | 0.40 sec | 2.05 sec |
| Before enhancement | Entities (Nodes) | 1,711 | 3,552 |
| | Triplets (Edges) | 6,155 | 18,936 |
| | Type of Relations | 24 | 25 |
| After enhancement | Entities (Nodes) | 1,711 | 3,552 |
| | Triplets (Edges) | 16,367 | 48,395 |
| | Type of Relations | 27 | 27 |

Table 4: Statistics of the knowledge graphs for the human relationships.

### 6.2.1 Questions

To evaluate the performance of the question-answering system, we collected 35 questions from rāmāyaṇa and 45 questions from mahābhārata from 12 different users, with each user contributing between 5-10 questions.

### 6.2.2 Performance

We evaluate the performance of the system for three tasks.

- **QParse** refers to the query parsing task. If the query pattern is correctly formed from the natural language question, we count it as a success; otherwise, it is a failure.

- **QCond** is the conditional question answering task subject to correct query formation. A success is counted only if the answer to the question is completely correct.

- **QAll** is the overall question answering task.

Table 5 demonstrates the performance of our system on the collected questions. The query parsing task is fairly accurate. However, the accuracy of question-answering has a lot of scope for improvement. We next analyze some of the reasons for failure.

## 6.3 Analysis of Wrong Answers

We analyze the wrong answers in two phases: parsing errors and answering errors.

| Text | Task | Total | Found | Correct | Precision | Recall | F1 |
|---|---|---|---|---|---|---|---|
| rāmāyaṇa | QParse | 35 | 33 | 27 | 0.82 | 0.77 | 0.79 |
| | QCond | 27 | 19 | 09 | 0.47 | 0.33 | 0.39 |
| | QAll | 35 | 20 | 10 | 0.50 | 0.29 | 0.37 |
| mahābhārata | QParse | 45 | 45 | 41 | 0.91 | 0.91 | 0.91 |
| | QCond | 41 | 36 | 22 | 0.61 | 0.54 | 0.57 |
| | QAll | 45 | 40 | 23 | 0.58 | 0.51 | 0.54 |
| Combined | QParse | 80 | 78 | 68 | 0.87 | 0.85 | 0.86 |
| | QCond | 60 | 55 | 31 | 0.56 | 0.46 | 0.50 |
| | QAll | 80 | 60 | 33 | 0.55 | 0.41 | 0.47 |

Table 5: Performance of the question-answering tasks.

### 6.3.1 Parsing Errors

Following are some examples of queries that got incorrectly parsed.

- गान्धार्याः पुत्राणाम् नामानि कानि → [गान्धारी, पुत्र, किम्]
  The question expects all the names of sons of gāndhārī गान्धारी but the parsed query only asks for the name of 'a son' of गान्धारी. This error originates from the fact that we have not considered the number (वचन) of the relationship word while parsing the question. Strictly speaking, however, the question is not a simple factoid question. Nevertheless, number (वचन) can be considered, and all triplets that satisfy the criteria can be returned.

- कर्णार्जुनयोः कः सम्बन्धः → [किम्, किम्, सम्बन्ध]
  There are patterns in the question set that are not handled by our algorithm. For example, the algorithm did not handle the way of asking the relationship between two people using the word सम्बन्ध and, thus, results in a triplet that does not make sense. If the same question was phrased as कर्णः अर्जुनस्य कः, our algorithm would be able to parse the question to give [अर्जुन, किम्, कर्ण]. Questions like कर्णः अर्जुनस्य कः, अर्जुनस्य कर्णः कः, अर्जुनस्य कः कर्णः and कर्णः कः अर्जुनस्य also get parsed correctly to [अर्जुन, किम्, कर्ण].

- विवाहः अर्जुनस्य अभवत् कया सह → [अर्जुन, किम्, विवाह]
  The question parsing algorithm, while tolerant to some extent, is not fully robust to free word order. An occurrence of विवाह word needs to be followed by the instrumental case (तृतीया) word, followed by सह for it to be parsed correctly. Thus, if the question is changed to अर्जुनस्य विवाहः कया सह अभवत्, it will get parsed correctly to yield [अर्जुन, पत्नी, किम्].

### 6.3.2 Answering Errors

Out of the queries that correctly get parsed, following are the queries which we cannot find the answer due to the inability of performing path queries.

- ऊर्मिला दशरथस्य का → [दशरथ, किम्, ऊर्मिला]
  This question would have got answered only if there is a direct edge between दशरथ and ऊर्मिला. If there is no direct edge, but an edge between दशरथ and लक्ष्मण exists along with the edge between लक्ष्मण and ऊर्मिला, then this answer should have been found. Our inability to pose it as a graph path searching query is the cause of this failure.

- हनुमतः पिता कः → [हनुमत्, पितृ, किम्]
  We correctly parse this question and there exists a triplet [मारुति, पितृ, पवन]. However, as the information that मारुति is another name of हनुमत् is not present in the knowledge graph, resulting in the failure to answer this question.

| śloka | sandhi-samāsa **split** |
|---|---|
| अनिलस्य शिवा भार्या तस्याः पुत्रो मनोजवः। | अनिलस्य शिवा भार्या तस्याः पुत्रः मनोजवः। |
| अविज्ञातगतिश्चैव द्वौ पुत्रावनिलस्य तु॥२५॥ | अविज्ञात-गतिः-च-एव द्वौ पुत्रौ=-अनिलस्य तु॥२५॥ |
| प्रत्यूषस्य विदुः पुत्रमृषिं नाम्नाऽथ देवलम्। | प्रत्यूषस्य विदुः पुत्रम्-ऋषिम् नाम्ना-अथ देवलम्। |
| द्वौ पुत्रौ देवलस्यापि क्षमावन्तौ मनीषिणौ। | द्वौ पुत्रौ देवलस्य-अपि क्षमावन्तौ मनीषिणौ। |
| बृहस्पतेस्तु भगिनी वरस्त्री ब्रह्मवादिनी॥२६॥ | बृहस्पतेः-तु भगिनी वर-स्त्री ब्रह्म-वादिनी॥२६॥ |
| योगसिद्धा जगत्कृत्स्नमसक्ता विचचार ह। | योग-सिद्धाः जगत्-कृत्स्नम्-असक्ता विचचार ह। |
| प्रभासस्य तु भार्या सा वसूनामष्टमस्य ह॥२७॥ | प्रभासस्य तु भार्या सा वसूनाम्-अष्टमस्य ह॥२७॥ |

Table 6: śloka 25, 26, 27 from adhyāya 67 of ādi parvan in mahābhārata.

- पुरोः कः वंशजः यस्य पुत्रः अर्जुनः → [पुरु, वंशज, किम्], [यद्, पुत्र, अर्जुन]
  Again, despite getting correctly parsed, since we cannot follow the "has-son" relationship arbitrary number of times, this query cannot be answered.

### 6.3.3 Correct Answers despite Wrong Parsing

Interestingly, there are cases when despite the query being parsed incorrectly, the correct answer exists in the result set. The following examples highlight two such cases.

- रावणस्य कनिष्ठतमः भ्राता कः → [रावण, भ्रातृ, किम्]
  The triplet is incorrectly formed, since we did not capture the information कनिष्ठतमः (youngest). However, the correct answer, विभीषण, being a brother of रावण, is captured in the result set. The question is, thus, deemed to be answered correctly.

- भीमस्य अग्रजः कः आसीत् → [भीम, भ्रातृ, किम्]
  Similar to the previous question, we classify the formed triplet as incorrect, for missing the quality 'elder'. However, answers found do contain the correct answers युधिष्ठिर and कर्ण.

## 6.4 Analysis of Errors in KG Triplets

We now take a look at in-depth analysis of some incorrect triplets retrieved by our method and investigate the reasons behind the failure. For this purpose, we consider a small extract from the corpus and follow the entire pipeline of forming the triplets.

Table 6 gives an extract containing three śloka (25, 26 and 27) from adhyāya 67 of the ādi parvan in mahābhārata. Table 7, Table 8 and Table 9 contain the detailed analysis of these śloka as well as a classification of the errors in the analysis.

### 6.4.1 Types of Errors

We now discuss the possible errors, as exemplified in the analysis tables 7, 8 and 9.

- **AnalysisError**:
  This is an error in the analysis obtained from *The Sanskrit Heritage Parser*. For example, the word भार्या in śloka 25 is analysed as a form of भारि instead of a form of भार्या. Thus, the prātipadika identified is wrong. This also results in the other analysis details such as case, gender and number, being wrong. It should be noted that words can be analyzed differently in different contexts. For example, the word भार्या, if analyzed standalone as a word, can get analyzed correctly; however, in the current context, it results in an erroneous analysis.[16]

- **OversplitError**:
  This is an error in the sandhi and samāsa splitter, where a word that should not have been split is split. For example, in śloka 26, वरस्त्री is wrongly oversplit as वर and स्त्री, and ब्रह्मवादिनी as ब्रह्म and वादिन्. Sometimes a word is erroneously oversplit by the analyser as well. Again, in śloka 26, for example, वादिन् is erroneously split as वा and आदिन्.

---
[16]Erroneous analysis of भार्या: https://sanskrit.inria.fr/cgi-bin/SKT/sktreader.cgi?lex=SH&st=t&us=f&cp=t&text=anilasya+zivaa+bhaaryaa+tasyaa.h+putra.h+manojava.h&t=VH&mode=p

| Word | Root | Analysis | Is-Noun | Is-Verb | Error |
|---|---|---|---|---|---|
| अनिलस्य | अनिल | ['g.', 'sg.', 'm.'] | True | False | |
| शिवा | शिव | ['nom.', 'sg.', 'f.'] | True | False | |
| भार्या | भारि | ['i.', 'sg.', 'f.'] | True | False | AnalysisError |
| तस्याः | तद् | ['g.', 'sg.', 'f.'] | False | False | |
| पुत्रः | पुत्र | ['nom.', 'sg.', 'm.'] | True | False | |
| मनो जवः | मनोजव | ['nom.', 'sg.', 'm.'] | True | False | Corrected |
| अविज्ञा | अविज्ञ | ['nom.', 'sg.', 'f.'] | True | False | OversplitError |
| आत | अत् | ['pft.', 'ac.', 'pl.', '2'] | False | True | OversplitError |
| गतिः | गति | ['nom.', 'sg.', 'f.'] | True | False | OversplitError |
| च | च | ['conj.'] | False | False | |
| एव | एव | ['prep.'] | False | False | |
| द्वौ | द्व | ['acc.', 'du.', 'm.'] | True | False | |
| पुत्रौ | पुत्र | ['acc.', 'du.', 'm.'] | True | False | |
| अनिलस्य | अनिल | ['g.', 'sg.', 'm.'] | True | False | |
| तु | तु | ['conj.'] | False | False | |

Table 7: Analysis of śloka 25.

- **SandhiSamaasaError**:
  There can be error in analyzing the correct sandhi and samāsa in a word. In other words, when a word is broken, the constituent words can be erroneous. For example, in śloka 27, योगसिद्धा जगत् is split as योग, सिद्धाः and जगत्, where योगसिद्धा, in addition to being oversplit, is also changed into plural form.

### 6.4.2 Extracting Triplets

After obtaining the analysis, when we proceed to extract triplets as mentioned, we tried using 4 different filters for extracting triplets. In every filter, the case of the subject word must be sixth (षष्ठी) and the gender of the object word must match with the gender of the predicate word. Filters differ in the allowed positions of subject and object words relative to the predicate word as well whether the number (वचन) of the object is matched or not.

Table 10 describe the different filters. Filter 1 is the superset of other filters and Filter 2 is the superset of Filter 3 and Filter 4.

Through empirical evidence, we found that Filter 2, although being stricter than Filter 1, still captures roughly the same number of triplets while reducing the errors. Filter 3 and Filter 4, while exhibiting fewer mistakes, find fewer correct triplets as well. While we acknowledge that such an analysis is required on a larger scale to decide among the filters, for our purposes, we choose Filter 2 based on the empirical evidence, and proceed further.

### 6.4.3 Analysis of Incorrect Triplets

In this section, we take a look at some wrong triplets that were retrieved and the reasons behind their retrieval.

- (प्रत्यूष, पुत्र, मनीषिन्)
  śloka 26, listed in Table 6 contains two relationship words, पुत्रम् and पुत्रौ. The first one is used in relation to देवल who is the son of प्रत्यूष, and the triplet (प्रत्यूष, पुत्र, देवल) is found correctly. However, because of the presence of the second word पुत्रौ, which is actually used with देवलस्य, a wrong triplet (प्रत्यूष, पुत्र, मनीषिन्) is formed. Due to the same reason, (प्रत्यूष, पुत्र, क्षमावत्) is also found. Since the context for finding relationships covers the full śloka, when a single śloka contain multiple relationships, such errors occur. If sentences were instead used, the error could have been reduced. However, there do not exist clear sentence boundaries.

| Word | Root | Analysis | Is-Noun | Is-Verb | Error |
|---|---|---|---|---|---|
| प्रत्यूषस्य | प्रत्यूष | ['g.', 'sg.', 'm.'] | True | False | |
| विदुः | विद् | ['pft.', 'ac.', 'pl.', '3'] | False | True | |
| पुत्रम् | पुत्र | ['acc.', 'sg.', 'm.'] | True | False | |
| ऋषिम् | ऋषि | ['acc.', 'sg.', 'm.'] | True | False | |
| नाम्ना | नामन् | ['adv.'] | False | False | |
| अथ | अथ | ['conj.'] | False | False | |
| देवलम् | देवल | ['acc.', 'sg.', 'm.'] | True | False | |
| द्वौ | द्व | ['acc.', 'du.', 'm.'] | True | False | |
| पुत्रौ | पुत्र | ['acc.', 'du.', 'm.'] | True | False | |
| देवलस्य | देवल | ['g.', 'sg.', 'm.'] | True | False | |
| अपि | अपि | ['conj.'] | False | False | |
| क्षमावन्तौ | क्षमावत् | ['acc.', 'du.', 'm.'] | True | False | |
| मनीषिणौ | मनीषिन् | ['acc.', 'du.', 'm.'] | True | False | |
| बृहस्पतेः | बृहस्पति | ['g.', 'sg.', 'm.'] | True | False | |
| तु | तु | ['conj.'] | False | False | |
| भगिनी | भगिनी | ['nom.', 'sg.', 'f.'] | True | False | |
| वर | वर | ['voc.', 'sg.', 'm.'] | True | False | OversplitError |
| स्त्री | स्त्री | ['nom.', 'sg.', 'f.'] | True | False | OversplitError |
| ब्रह्म | ब्रह्मन् | ['acc.', 'sg.', 'n.'] | True | False | OversplitError |
| वा | वा | ['conj.'] | False | False | OversplitError |
| आदिनी | आदिन् | ['acc.', 'du.', 'n.'] | True | False | OversplitError |

Table 8: Analysis of śloka 26.

| Word | Root | Analysis | Is-Noun | Is-Verb | Error |
|---|---|---|---|---|---|
| योग | योग | ['voc.', 'sg.', 'm.'] | True | False | OversplitError, AnalysisError |
| सिद्धाः | सिद्ध | ['acc.', 'pl.', 'f.'] | True | False | OversplitError, SandhiSamaasaError |
| जगत् | जगत् | ['acc.', 'sg.', 'n.'] | True | False | |
| कृत्स्नम् | कृत्स्न | ['acc.', 'sg.', 'm.'] | True | False | |
| असक्ता | असक्त | ['nom.', 'sg.', 'f.'] | True | False | |
| विचचार | वि-चर् | ['pft.', 'ac.', 'sg.', '3'] | False | True | |
| ह | ह | ['part.'] | False | False | |
| प्रभासस्य | प्रभास | ['g.', 'sg.', 'm.'] | True | False | |
| तु | तु | ['conj.'] | False | False | |
| भार्या | भार्य | ['nom.', 'sg.', 'f.'] | True | False | |
| सा | तद् | ['nom.', 'sg.', 'f.'] | False | False | |
| वसूनाम् | वसु | ['g.', 'pl.', 'm.'] | True | False | |
| अष्टमस्य | अष्टम | ['g.', 'sg.', 'm.'] | True | False | |
| ह | ह | ['part.'] | False | False | |

Table 9: Analysis of śloka 27.

| Filter | Position of subject | Position of object | Number (वचन) of object |
|---|---|---|---|
| 1 | Either side of predicate | Either side of predicate | Does not matter |
| 2 | Either side of predicate | Either side of predicate | Must match predicate |
| 3 | Before predicate | After predicate | Must match predicate |
| 4 | After predicate | Before predicate | Must match predicate |

Table 10: Filters for extracting triplets.

| Scenario | Training Set | Testing Set |
|---|---|---|
| S1 | First 20% of adhyāya 1 | Rest 80% of adhyāya 1 |
| S2 | First 20% of adhyāya 2 | Rest 80% of adhyāya 2 |
| S3 | adhyāya 1 | adhyāya 2 |
| S4 | adhyāya 2 | adhyāya 1 |

Table 11: Training and testing scenarios on bhāvaprakāśa nighaṇṭu.

- (बृहस्पति, भगिनी, स्त्री)
  As discussed in Section 6.4.1, the word वरस्त्री gets oversplit wrongly into वर and स्त्री, and the split words are analysed separately, resulting in the wrong triplet. Even if this split did not occur, we would have got वरस्त्री as the object in this triplet. This is wrong since this is actually an adjective used for the sister of बृहस्पति. Since we currently do not have any mechanism of distinguishing between nouns and adjectives, it would have resulted in incorrect triplets.

We next examine some triplets that should have been found but were not found and the reasons behind their non-retrieval.

- (अनिल, पत्नी, शिवा)
  The relationship word that occurs in śloka 25 in Table 6 is भार्या, which suffers an AnalysisError and is identified as तृतीया of भारि instead of प्रथमा of भार्या. Due to the root word (प्रातिपदिक) itself being misidentified, it is not recognized as a relationship word and thus, does not satisfy the filtering criterion. Consequently, the triplet (अनिल, पत्नी, शिवा) is missed.

- (प्रभास, पत्नी, ब्रह्मवादिनी)
  In śloka 27, भार्या of प्रभास is referred to with a pronoun सा, which is connected to a noun in the previous śloka. To correctly identify the triplet (प्रभास, पत्नी, ब्रह्मवादिनी), we would need a mechanism to connect pronouns to their proper subjects. We do not handle this currently.

## 6.5 Synonym Identification from bhāvaprakāśa nighaṇṭu

Questions for the bhāvaprakāśa are implicit, as we are considering only the synonymous relationship. Therefore, the evaluation is performed on the *synonym groups* and *synonym pairs* identification. We created ground truth for the first two adhyāya of bhāvaprakāśa nighaṇṭu. adhyāya 1 contains 261 śloka, while adhyāya 2 contains 131 śloka. For each of these śloka, we first identified if it is a *synonym* śloka. If it is so, we next extracted the list of synonymous words contained in it.

### 6.5.1 Classification

Using the feature vectors obtained for each śloka, we used various classifiers to classify each śloka as a synonym śloka or otherwise. We tried four practical scenarios of training and testing set choices as described in Table 11.

| Scenario | Train Size | Test Size | $P$ | $P'$ | $TP$ | Accuracy | Precision | Recall | F1 |
|---|---|---|---|---|---|---|---|---|---|
| S1 | 52 | 209 | 84 | 56 | 42 | 0.73 | 0.75 | 0.50 | 0.60 |
| S2 | 26 | 105 | 44 | 43 | 31 | 0.76 | 0.72 | 0.71 | 0.71 |
| S3 | 261 | 131 | 54 | 45 | 36 | 0.79 | 0.80 | 0.67 | 0.73 |
| S4 | 131 | 261 | 90 | 99 | 66 | 0.78 | 0.67 | 0.73 | 0.70 |

Table 12: Performance of classifiers in identifying synonym śloka.

| False Positives (9) | False Negatives (18) |
|---|---|
| कामरूपोद्भवा कृष्णा नैपाली नीलवर्णयुक् काश्मीरी कपिलच्छाया कस्तूरी त्रिविधा स्मृता ॥६॥ | श्रीखण्डं चन्दनं न स्त्री भद्र श्रीस्तैलपर्णिकः गन्धसारो मलजयस्तथा चन्द्र द्युतिश्च सः ॥११॥ |
| महिषाक्षो महानीलः कुमुदः पद्म इत्यपि हिरण्यः पद्ममो ज्ञेयो गुग्गुलोः पञ्च जातयः ॥३३॥ | भद्र मुस्तञ्च गुन्द्रा च तथा नागरमुस्तकः मुस्तं कटु हिमं ग्राहि तिक्तं दीपनपाचनम् ॥९३॥ |

Table 13: Examples of errors in classification (scenario S3).

The size of training sets were chosen to be smaller than those of test sets to resemble the real-world scenario where the ground truth can be created for only a small portion of the text, and predictions are needed to be made on the rest.

Table 12 shows the performance of the best classifier under various scenarios in identifying the śloka containing synonyms.

Table 13 shows some examples of wrongly classified śloka for the best performing scenario S3.

### 6.5.2 Synonym Identification

We next evaluate the performance of finding synonymous pairs from a synonym śloka. Table 14 shows the performance in identifying groups of synonymous substances. We say that a group of substances is *covered* even if a single pair in the group is identified. The result shows that even this has a scope for improvement.

Table 15 shows an example of a synonym śloka where none of the pairs are extracted correctly. The correct synonyms are चन्द्रिका, चर्महन्त्री, पशुमेहनकारिका, नन्दिनी, कारवी, भद्रा, वासपुष्पा, सुवासरा. We find the pairs (कारिका, हन्त्री), (कारिका, भद्र), (कारिका, सपुष्प), (नन्दिन, रवि), (भद्र, हन्त्री), (भद्र, सपुष्प), (सपुष्प, हन्त्री), none of which are correct. The reasons for the errors are shown in Table 16. Almost all the nouns are analysed incorrectly, resulting in the group being completely missed.

In addition to the errors discussed in Section 6.4.1, an additional error occurs here, that of **TextError**. This refers to an error in the text corpus that we are working with. In particular, the original śloka contains the word चन्द्रिका while the corpus we are working with, has that word split as चन्द्रि and का, which results in this word not being analysed correctly. After correcting this error manually, we now obtain a valid pair (चन्द्रिका, भद्रा), thus covering this group.

We next analyse the finer errors that occur when some members of a synonymous group are identified correctly, but not all. Table 17 shows the performance.

Table 18 shows a synonym śloka from adhyāya 1 (हरीतक्यादिवर्गः).

This śloka contains a total of 11 synonyms. We find pairs of synonyms involving 9 out of these, synonym pairs involving 8 of which are correct. We show examples of some of the false negatives and false positives among the pairs of synonyms identified.

| | Synonym śloka | Groups present | Groups found | Group coverage |
|---|---|---|---|---|
| adhyāya 1 | 90 | 87 | 60 | 0.69 |
| adhyāya 2 | 54 | 53 | 39 | 0.74 |

Table 14: Group coverage in synonym pair identification.

| Synonym śloka | sandhi-samāsa split |
|---|---|
| चन्द्रि का चर्महन्त्री च पशुमेहनकारिका। | चन्द्रि का चर्महन्त्री च पशुमेहन-कारिका। |
| नन्दिनी कारवी भद्रा वासपुष्या सुवासरा ॥९६॥ | नन्दिनी कारवी भद्रा वासपुष्या सु-वासराः ॥९६॥ |

Table 15: śloka 96 from adhyāya 1 of bhāvaprakāśanighaṇṭu and its sandhi-samāsa split.

| Word | Root | Analysis | Is-Noun | Is-Verb | Error |
|---|---|---|---|---|---|
| चन्द्रि | चन्द्रि | ['?'] | False | False | TextError |
| का | किम् | ['nom.', 'sg.', 'f.'] | False | False | TextError |
| चर्म | चर्मन् | ['acc.', 'sg.', 'n.'] | True | False | OversplitError |
| हन्त्री | हन्तृ | ['nom.', 'sg.', 'f.'] | True | False | OversplitError |
| च | च | ['conj.'] | False | False | |
| पशुमेहन | पशुमेहन | ['voc.', 'sg.', 'n.'] | True | False | OversplitError |
| कारिका | कारिका | ['nom.', 'sg.', 'f.'] | True | False | OversplitError |
| नन्दिनी | नन्दिन् | ['acc.', 'du.', 'n.'] | True | False | AnalysisError |
| का | किम् | ['nom.', 'sg.', 'f.'] | False | False | OversplitError |
| रवी | रवि | ['acc.', 'du.', 'm.'] | True | False | OversplitError |
| भद्रा | भद्र | ['nom.', 'sg.', 'f.'] | True | False | |
| वा | वा | ['conj.'] | False | False | OversplitError |
| सपुष्या | सपुष्य | ['nom.', 'sg.', 'f.'] | True | False | OversplitError |
| सु | सु | ['?'] | False | False | OversplitError |
| वासराः | वासर | ['voc.', 'pl.', 'm.'] | True | False | OversplitError |

Table 16: Analysis of śloka 96.

- **False Positive:** (अमृता, अवी)
  The word अव्यथा is split wrongly as अवी and अथा, and are then analysed separately. This results in both अमृता and अवी being in the same case (प्रथमा) and same number (एकवचन), thus getting wrongly marked as a synonymous pair.

- **False Negative:** (अभया, अमृता)
  The word अभया gets analysed as instrumental (तृतीया) case of अभा instead of nominative (प्रथमा) case of अभया. This results in a case mismatch with अमृता and the pair is not extracted as a synonymous pair.

# 7 Conclusions and Future Work

In this paper, we have designed a framework to build a knowledge graph (KG) directly from saṃskṛta texts, and use it for question-answering in saṃskṛta. Our framework has multiple components and is based on rules and heuristics developed using the knowledge of grammar of saṃskṛta language and structure of the text.

However, for almost all the components, the accuracy can be improved. Improvements on any of these components by us or by others will make the system better. In future, we would like to work on improving the modules in a systematic manner. The biggest source of improvement can possibly come from a better word analyser. Usage of dictionaries, thesauri (such as amarakośa) and Sanskrit WordNet will be explored to see if they can help in understanding the structure of a word better. Crowd sourcing tools as well as human experts can also help refine some of the steps. We would also like to expand the question-answering framework to work with longer questions that are not necessarily of the type factoid.

|           | śloka | **Synonym** śloka | $P$ | $P'$ | $TP$ | **Precision** | **Recall** | **F1** |
|-----------|-------|-------------------|-----|------|------|---------------|------------|--------|
| adhyāya 1 | 231   | 90                | 534 | 562  | 369  | 0.66          | 0.69       | 0.67   |
| adhyāya 2 | 161   | 54                | 300 | 348  | 214  | 0.62          | 0.71       | 0.66   |

Table 17: Performance of finding synonym pairs.

| **Synonym** śloka | sandhi-samāsa **split** | $P$ | $P'$ | $TP$ |
|-------------------|------------------------|-----|------|------|
| हरीतक्यभया पथ्या कायस्था पूतनाऽमृता हैमवत्यव्यथा चापि चेतकी श्रेयसी शिवाः ॥६॥ | हरीतकी-अभया पथ्या कायस्था पूतना-अमृता हैमवती-अव्यथा च-अपि चेतकी श्रेयसी शिवाः ॥६॥ | 11 | 9 | 8 |

Table 18: Example of wrong pairs from adhyāya 1 of bhāvaprakāśa nighaṇṭu.

To conclude, we hope that this effort serves as a step towards the ultimate aim of automatically building a full-fledged knowledge graph from a saṃskṛta corpus.

## Acknowledgements

We thank Dr. Sai Susarla, Dean at Maharshi Veda Vyas MIT School of Vedic Sciences, Pune, India, and his team, for sharing their expertise in āyurveda with us. We thank Shubhangi Agarwal and Rujuta Pimprikar for the help in creating ground truth as well as providing valuable feedback from time to time. We thank Dr. Kripabandhu Ghosh and Garima Gaur for the discussions and valuable feedback. We thank our saṃskṛta teacher Pralay Manna for enabling us in understanding the language better. We also thank the anonymous reviewers for their comments and suggestions.